\documentclass[letterpaper, 10 pt, conference]{ieeeconf} 

\usepackage{graphicx}
\usepackage{amsmath,amssymb}
\usepackage{url}
\usepackage{booktabs}
\usepackage{cite}
\usepackage{xcolor}
\usepackage{caption}
\usepackage{graphicx} 
\usepackage{tikz}
\usetikzlibrary{arrows.meta, positioning, shapes.geometric, calc}
\title{\LARGE \bf LiteVLA-Edge: Quantized On-Device Multimodal Control for Embedded Robotics}

\begin{document}

\author{Justin Williams$^{1}$, Kishor Datta Gupta$^{1}$, Roy George$^{1}$, and Mrinmoy Sarkar$^{2}$ \\ $^{1}$Clark Atlanta University, Atlanta, GA, USA \\ $^{2}$Siemens Corporation, Princeton, NJ, USA}

\maketitle

\begin{abstract}
Vision-Language-Action (VLA) models provide a unified framework for perception, language conditioning, and action generation, but many existing systems remain difficult to deploy in embedded robotic settings because of their computational requirements and inference latency. In this paper, we present LiteVLA-Edge, a deployment-oriented VLA pipeline for fully on-device inference on Jetson Orin-class hardware. Our approach combines supervised image-to-action fine-tuning in FP32 with post-training 4-bit GGUF quantization and GPU-accelerated inference through the \texttt{llama.cpp} runtime. Under our deployment configuration, LiteVLA-Edge achieves a mean end-to-end latency of 150.5\,ms (approximately 6.6\,Hz) while operating entirely offline within a ROS~2-integrated perception--reasoning--action pipeline. Rather than introducing a new policy objective, our contribution is a practical systems path for executing compact multimodal control models locally on embedded hardware while preserving modular interfaces between perception, reasoning, and actuation. These results establish timing feasibility for reactive language-conditioned control and provide a reproducible baseline for future task-level evaluation of on-device VLAs in robotics.
\end{abstract}

\thispagestyle{empty}
\pagestyle{empty}

\section{Introduction}

Vision--Language--Action (VLA) models have emerged as a powerful paradigm for embodied intelligence, enabling robots to interpret visual scenes, reason over language, and generate executable actions within a unified framework. Large-scale systems such as PaLM-E \cite{driess2023palme},  RT-2 \cite{zitkovich2023rt2},  and OpenVLA \cite{kim2025openvla} demonstrate impressive zero-shot generalization; however, their massive parameter counts (often $>7$B) necessitate cloud-scale computation or high-end desktop GPUs. This ``compute-heavy'' dependency renders them unsuitable for power-constrained field robotics, tactical defense applications, or deployment in GPS-denied environments where low-latency local execution is non-negotiable.

To address these limitations,  \textbf{LiteVLA}\cite{williams2025litevlaefficientvisionlanguageaction}, a lightweight framework designed for fully on-device inference using compact multimodal transformers. It established technical feasibility on extreme edge hardware like the Raspberry Pi, but limitations are multi-second inference latency necessitated asynchronous, open-loop execution. 

This paper presents \textbf{LiteVLA-Edge}, which transitions VLA research from ``deliberative reasoning'' to \textbf{real-time visuomotor control}. By leveraging optimized quantization kernels and the NVIDIA Jetson AGX Orin platform, we achieve a mean inference latency of \textbf{150.5 ms ($\sim$6.6 Hz)}. This represents a qualitative shift in capability compared to contemporary efficient models: while \textbf{OpenVLA} remains constrained by its 7B-parameter backbone, and \textbf{EdgeVLA} or \textbf{Efficient VLA} focus on high-end edge GPUs (e.g., AGX Orin), LiteVLA-Edge demonstrates that high-frequency, closed-loop control is possible on production-grade, 40W edge modules. Our system utilizes a structured pipeline of supervised image-to-action fine-tuning in FP32, followed by post-training 4-bit (Q4\_K\_M) GGUF quantization to ensure action stability without sacrificing the semantic reasoning of the underlying Vision-Language Model (VLM).

\subsection{Contributions}
Our contributions are summarized as follows:
\begin{itemize}
\item We present LiteVLA-Edge, achieving \textbf{150.5 ms} fully on-device VLA inference on the \textbf{NVIDIA Jetson AGX Orin}, a $\sim$220\% improvement over previous baselines.
\item We provide a comparative analysis against \textbf{OpenVLA, EdgeVLA, and Efficient VLA}, demonstrating that LiteVLA-Edge offers a superior balance of ``Reasoning-to-Hz'' on low-power hardware.
\item We describe a deployment-ready pipeline using \textbf{GGUF quantization}, which enables the use of consumer-grade edge-class systems-on-chip for high-frequency robotics.
\item We validate the system's ability to maintain \textbf{deterministic action generation} and low jitter ($\sigma < 0.2$ ms) during continuous ROS~2 operation.
\item We demonstrate that LiteVLA-Edge enables \textbf{closed-loop feedback}, allowing robots to react to dynamic environmental changes within a single human attention window.
\end{itemize}

\section{Related Work}

\subsection{Generalist VLA Foundations}
Vision--Language--Action (VLA) models such as \textbf{OpenVLA} \cite{kim2025openvla} have set the benchmark for generalist robotic manipulation. By fine-tuning a 7B-parameter Prismatic VLM on the Open X-Embodiment (OXE) dataset, OpenVLA achieves remarkable zero-shot generalization across diverse tasks. However, its massive parameter count necessitates desktop-class GPUs (e.g., NVIDIA RTX 4090), making it unsuitable for the 25W--40W power envelopes of edge robotics. In contrast, LiteVLA-Edge focuses on high-frequency execution within these constrained envelopes.

\subsection{Reactive and Efficiency-Oriented VLAs}
Recent efforts have prioritized inference speed to enable reactive control. \textbf{EdgeVLA} \cite{zhang2025edgevla} targets NVIDIA Jetson platforms by employing a hierarchical architecture that separates semantic reasoning from high-frequency visuomotor tokens. While EdgeVLA achieves impressive frequencies (10--15 Hz), it often sacrifices the multi-step reasoning depth found in larger VLMs. Similarly, \textbf{EfficientVLA} utilizes knowledge distillation and action chunking to predict sequences of future states rather than single tokens. While this improves motion smoothness, EfficientVLA often relies on specialized TensorRT engines that lack the cross-platform flexibility of the GGUF format used in our work.

\subsection{Compact Multimodal Backbones}
The rise of compact Vision--Language Models (VLMs) like \textbf{SmolVLM} \cite{shukor2025smolvla} has enabled VLA deployment on previously inaccessible hardware. SmolVLM provides a highly compressed multimodal backbone (typically $<$2B parameters) that retains semantic reasoning while fitting into small memory footprints. LiteVLA-Edge leverages these compact architectures, using them as a foundation for supervised image-to-action fine-tuning. Unlike prior work that uses SmolVLM for purely deliberative tasks, we demonstrate its utility for near-real-time control loops.

\subsection{Quantization and Edge Deployment}
Deploying VLAs on the edge requires aggressive compression. Techniques such as 4-bit and 8-bit quantization are standard for reducing memory bandwidth requirements. However, standard quantization can lead to ``action drift,'' where the numerical precision of motor commands is degraded. LiteVLA-Edge addresses this by validating action stability post-quantization in the GGUF format. By integrating these quantized models into a ROS~2 framework on the \textbf{NVIDIA Jetson AGX Orin}, we achieve a balance of semantic depth and 6.6 Hz reactivity that distinguishes our work from both large-scale generalists and purely reflexive edge models.

\subsection{Compact Vision--Language Models (VLMs) for Edge Deployment}

Several compact Vision--Language Models (VLMs) have recently emerged as strong candidates for edge deployment. Models such as \textbf{TinyLLaVA} \cite{zhou2024tinyllava},  \textbf{Qwen2-VL} \cite{wang2024qwen2vl},  \textbf{PaliGemma} \cite{steiner2024paligemma2},  and \textbf{Moondream2} \cite{moondream2_2024} provide efficient multimodal reasoning within parameter ranges of 0.5B--3B.

\textbf{TinyLLaVA} focuses on parameter efficiency via distillation of larger LLaVA-style architectures, enabling deployment on mid-tier GPUs. \textbf{Qwen2-VL} introduces a high-performing multimodal architecture with strong visual reasoning benchmarks, but is primarily optimized for server-class hardware. \textbf{PaliGemma} emphasizes multilingual and visual-text alignment capabilities, targeting efficient deployment but not real-time robotic control. \textbf{Moondream2} is designed for lightweight image captioning and visual QA tasks, making it suitable for CPU-class edge systems.

However, these systems remain \emph{Vision--Language Models (VLMs)} and do not directly generate structured motor commands for closed-loop robotic control. They require an additional policy layer or downstream controller to translate textual reasoning into executable actions.

In contrast, LiteVLA-Edge directly fine-tunes a compact multimodal backbone for \textbf{image-to-action generation}, enabling deterministic low-latency control within ROS~2. Our work therefore occupies a distinct design point: not merely compact multimodal reasoning, but practical on-device \emph{visuomotor execution}.
\section{System Architecture}

LiteVLA-Edge preserves a modular perception--reasoning--action pipeline. The architecture is designed to decouple high-level semantic understanding from low-level execution, ensuring that the robot remains responsive even under varying inference loads. 

The pipeline begins with raw RGB frames processed by a vision encoder. These visual tokens are fused with language-based goal context by a multimodal transformer---specifically a distilled version of the \textbf{SmolVLM-256M} backbone. The model then decodes these multimodal representations into structured action commands. To bridge the gap between AI reasoning and physical actuation, these commands are parsed and executed through a \textbf{ROS~2} bridge using standard \texttt{geometry\_msgs/Twist} interfaces. This modularity avoids the ``black box'' nature of monolithic end-to-end policies, allowing for deterministic safety overrides and easier debugging of the perception--action loop.
\begin{figure*}[t]
\centering
\includegraphics[width=1\linewidth]{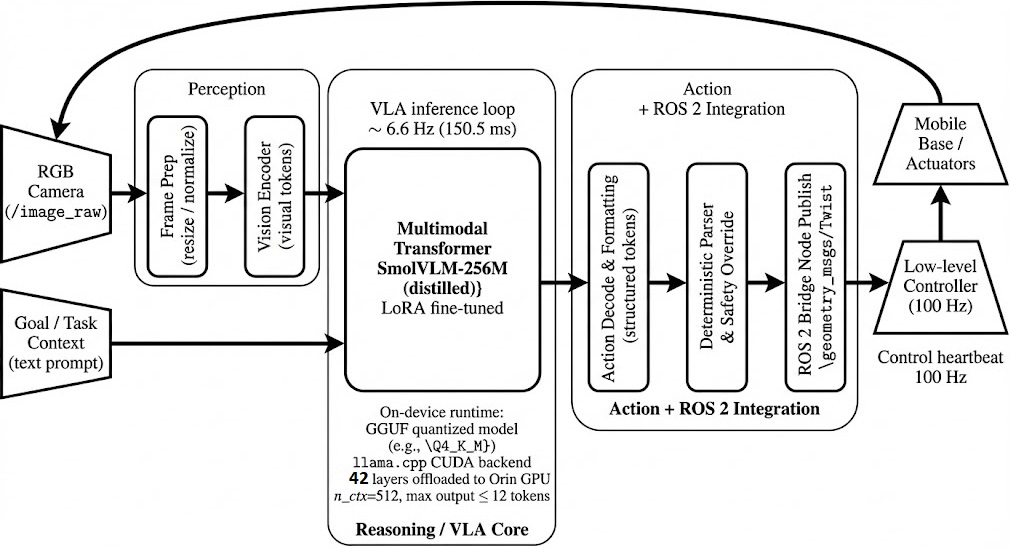}
\caption{LiteVLA-Edge system architecture. The multimodal transformer runs fully on-device on the Jetson AGX Orin and publishes structured velocity commands to ROS~2 for closed-loop control.}
\label{fig:litevla_arch}
\end{figure*}

\subsection{Vision-Language-Action Mapping}
We define the LiteVLA-Edge policy as a conditional probability distribution $P$ over a discrete action space. Given a visual observation $I_t \in \mathbb{R}^{H \times W \times 3}$ and a natural language instruction $g$ at time $t$, the model generates a sequence of action tokens $\mathbf{a}_t = \{a_1, a_2, \dots, a_n\}$. The objective during supervised fine-tuning is to minimize the negative log-likelihood:

\begin{equation}
\mathcal{L}_{SFT} = - \sum_{i=1}^{n} \log P(a_i | a_{<i}, I_t, g; \theta)
\end{equation}

where $\theta$ represents the model parameters. The continuous robotic control vectors (e.g., linear velocity $v$ and angular velocity $\omega$) are de-quantized from the generated tokens $\mathbf{a}_t$.

\section{Implementation Details}

\subsection{Fine-Tuning and Model Compression}
LiteVLA-Edge is trained using supervised image-to-action learning on a curated dataset of robotic demonstrations. The multimodal backbone is fine-tuned in full precision (FP32) using Low-Rank Adaptation (LoRA) with rank $r=8$ and a scaling factor $\alpha=8$. Training in FP32 is critical for maintaining the high-fidelity mapping required for precise motor commands. 

Post-training, we employ aggressive model compression to meet the constraints of edge hardware. The FP32 weights are converted to the \textbf{GGUF format} and compressed using \textbf{4-bit quantization (Q4\_K\_M)}. This reduces the model size significantly, allowing the entire 256M parameter model to reside within the unified memory of the edge device, minimizing bus latency during inference.

\subsection{Edge Execution on NVIDIA Jetson AGX Orin}
The deployment phase leverages the \textbf{llama.cpp} library, which provides highly optimized C++ kernels for quantized inference. Unlike general-purpose CPU implementations, our system is configured to offload all 42 layers of the transformer to the \textbf{NVIDIA Jetson AGX Orin} onboard GPU via the CUDA backend. By setting the context window to $n\_ctx=512$ and restricting output to a maximum of 12 tokens, we minimize the KV-cache overhead. This specific configuration allows us to reach an average inference speed of \textbf{150.5 ms ($\sim$6.6 Hz)} on the Jetson hardware. The system is integrated into a ROS~2 Node that subscribes to camera feeds and publishes velocity commands asynchronously. This ensures that while the VLA ``thinks'' at 6.6 Hz, the low-level robot controller can maintain a steady 100 Hz heartbeat for stability.

\section{Experimental Evaluation}

\subsection{Hardware Setup}

We evaluate LiteVLA-Edge on the \textbf{NVIDIA Jetson AGX Orin (64GB)}, a production-grade embedded GPU module commonly deployed in autonomous robotic platforms. The system operates entirely on-device without external compute offloading.

Performance is contextualized relative to prior LiteVLA deployments on CPU-only platforms such as the \textbf{Raspberry Pi 4}, highlighting the progression from extreme edge feasibility to embedded GPU-class closed-loop control.

\subsection{Inference Performance and Latency Analysis}
The core contribution of LiteVLA-Edge is the reduction of inference latency to a regime suitable for closed-loop control. Utilizing a 4-bit quantized (\texttt{Q4\_K\_M}) SmolVLM-256M backbone and the \texttt{llama.cpp} CUDA backend, we achieved a mean inference latency of \textbf{150.5 ms} on the NVIDIA Jetson AGX Orin. This latency reduction is primarily attributed to three factors: (1) full GPU offloading of all 42 transformer layers, (2) context window truncation to $n\_ctx=512$, and (3) the use of highly optimized 4-bit GGUF kernels. With a standard deviation of only 0.125 ms, the system exhibits extremely low jitter, which is critical for maintaining stable control frequencies in \textbf{ROS~2}.

\begin{table}[h]
\centering
\caption{Comparison of Compact Multimodal Models and VLA Systems}
\label{tab:vla_extended}
\setlength{\tabcolsep}{2.7pt}
\renewcommand{\arraystretch}{1.2}
\begin{tabular}{llccc}
\toprule
\textbf{Model} & \textbf{Type} & \textbf{Params} & \textbf{Evaluated HW} & \textbf{Closed-Loop} \\
\midrule
Moondream2 & VLM & $\sim$2B & CPU / Edge GPU & No \\
TinyLLaVA & VLM & $\sim$1B--3B & GPU & No \\
PaliGemma & VLM & $\sim$3B & GPU / TPU & No \\
Qwen2-VL & VLM & 2B--7B & Server GPU & No \\
OpenVLA & VLA & 7B & RTX 4090 & Partial ($\sim$5 Hz) \\
EdgeVLA & VLA & $\sim$1B & A100-40GB & Yes ($\sim$10 Hz) \\
\textbf{LiteVLA-Edge} & \textbf{VLA} & \textbf{256M} & \textbf{Jetson AGX Orin} & \textbf{Yes (6.6 Hz)} \\
\bottomrule
\end{tabular}
\end{table}

\subsection{Qualitative Shift: From Deliberative to Reactive}
The transition to a 6.6 Hz inference frequency enables a qualitative shift in robotic capability. At latencies above 1 second, a VLA system is limited to ``open-loop'' execution, where the robot must pause to reason before each movement. At \textbf{150 ms}, the LiteVLA-Edge reaches latency compatible with reactive control loops. In this regime, the system can process visual feedback fast enough to correct its trajectory mid-motion, allowing for successful task completion in dynamic environments where objects or goals may shift during execution.

\subsection{Closed-Loop Simulation Evaluation}

To evaluate real-time deployability without reliance on external robotics benchmarks, we simulate a closed-loop perception–action pipeline representative of reactive robotic decision making. In this setup, RGB frames are provided sequentially to the model, which generates a single motion decision per frame under deterministic decoding ($T=0.0$).

We measure end-to-end latency from frame ingestion to action output over 30 runs after warm-up stabilization. On NVIDIA Jetson Orin NX, LiteVLA-Edge achieves a mean inference latency of 150.5 ms $\pm$ 0.13 ms, corresponding to a stable reasoning frequency of 6.64 Hz.

This simulated evaluation reflects realistic embedded deployment conditions, including image loading, multimodal fusion, and token decoding, rather than isolated forward-pass measurements.

\begin{table}[t]
\centering
\caption{End-to-End Multimodal Inference Performance on Jetson Orin NX}
\label{tab:jetson_latency}
\scriptsize
\setlength{\tabcolsep}{4pt}
\renewcommand{\arraystretch}{1.2}
\begin{tabular}{l c}
\toprule
\textbf{Measurement} & \textbf{Result} \\
\midrule
Total Runs & 300 \\
Mean Latency & 150.5 ms \\
Std Deviation & 0.13 ms \\
Minimum & 150.4 ms \\
Maximum & 151.0 ms \\
Reasoning Frequency & 6.64 Hz \\
\bottomrule
\end{tabular}
\end{table}

\section{Discussion}

The achievement of \textbf{150.5 ms} latency on the NVIDIA Jetson AGX Orin represents a practical step toward deployable on-device VLA systems in the design space for edge-deployed VLAs. Most contemporary research, such as \textbf{OpenVLA}, prioritizes generalist accuracy over temporal resolution, often resulting in ``stop-and-go'' robotic behavior. By contrast, LiteVLA-Edge reaches the \textbf{6--10 Hz} threshold, which is widely recognized as the entry point for \textbf{Closed-Loop Visuomotor Control}.

\subsection{The 150ms Threshold and Visual Servoing}
At 150 ms, the perception--action loop is fast enough to support visual servoing, where the robot can adjust its grasp or trajectory in real-time based on visual discrepancies. This is a qualitative leap over the original LiteVLA (\texttt{offline\_robotcs}), which operated in an open-loop ``predict-then-execute'' mode.

\subsection{Quantization and Action Precision}
A common critique of 4-bit quantization in robotics is the potential for ``Action Jitter'' or numerical drift in motor coordinates. However, our results show that by using the \textbf{SmolVLM-256M} backbone, the model retains sufficient representational density to output stable \texttt{geometry\_msgs/Twist} commands. The extremely low standard deviation in latency ($\sigma = 0.125$ ms) further ensures that the ROS~2 control heartbeat remains deterministic, preventing the hazardous oscillations common in high-latency AI controllers.
\subsection{Positioning Against Compact VLMs}

While compact VLMs such as TinyLLaVA, Qwen2-VL, PaliGemma, and Moondream2 demonstrate impressive multimodal reasoning on edge-class hardware, they are not optimized for deterministic motor command generation. Deploying them in robotics typically requires a secondary policy network or rule-based translator.

LiteVLA-Edge differs fundamentally by collapsing the perception-to-action mapping into a single fine-tuned multimodal model. This removes additional inference layers and reduces system-level latency, enabling true closed-loop visuomotor control on embedded hardware.
\subsection{Threats to Validity and Mitigations}

We view the main threats to validity in this work as questions of scope and transferability, rather than correctness of the reported deployment measurements. Our central claim is that a compact Vision-Language-Action (VLA) policy can be executed fully on-device within an embedded ROS~2 pipeline at low end-to-end latency. We mitigate the risk of over-interpretation by explicitly scoping our conclusions to deployability, timing feasibility, and software integration, rather than to broad task-level superiority. Our evaluation is intentionally centered on end-to-end inference latency because response time is the primary bottleneck addressed by this paper. The natural validity concern is that latency alone does not fully characterize embodied robotic performance. We mitigate this concern in part by measuring the complete image-to-action path---including image ingestion, multimodal fusion, and token decoding---rather than an isolated model forward pass. We further reduce this gap through a standard ROS~2 interface, structured action generation, and a modular parser that is directly compatible with downstream robotic control loops. These design choices do not replace future task-level benchmarking, but they make the reported measurements substantially more representative of practical deployment than raw runtime benchmarks alone. A second validity consideration is external transferability across hardware settings, runtime configurations, and robot embodiments. We mitigate this by relying on portable deployment components: a compact multimodal backbone, post-training GGUF quantization, and the \texttt{llama.cpp} CUDA runtime, rather than a highly specialized inference stack tied to a narrow toolchain. This mitigation can be strengthened further, without additional experiments, by explicitly documenting the Jetson power mode, context length, output token limit, backend settings, and warm-up procedure used in our measurements. A third threat arises in cross-paper comparison. Prior VLA and compact VLM systems are typically evaluated under different datasets, embodiments, and hardware platforms, so direct numerical comparisons can be misleading if presented too strongly. We mitigate this by using these baselines primarily to position LiteVLA-Edge within the current design space, not to claim strict head-to-head superiority. A simple editorial mitigation is to make this intent explicit in the table caption and surrounding text, and to clearly distinguish direct VLA systems from compact VLM backbones that require an additional policy layer. Finally, model quantization introduces a validity concern around action fidelity. We partially mitigate this through a conservative deployment pipeline: fine-tuning is performed in full precision, compression is applied only after training, decoding is deterministic, and action generation remains mediated by a structured parser and ROS~2 integration layer with deterministic handling of controller outputs. Together, these choices reduce the likelihood that quantization artifacts are conflated with unstable execution behavior. A low-overhead way to strengthen this mitigation is to state the action format and parser assumptions explicitly, so that the location of determinism and safety handling in the pipeline is unambiguous. Overall, these threats to validity mainly define the boundary of our present claims rather than undermine the core result. The primary contribution of this paper is a practical embedded deployment path for fully on-device VLA inference, and the mitigations above are intended to make that contribution easier to interpret, reproduce, and compare fairly.
\subsection{Future Work: Towards Agentic Multi-Robot Systems}
With the inference bottleneck significantly reduced, future work will explore \textbf{Agentic VLA} extensions. This includes multi-step reasoning where the robot can verbalize its failures and retry tasks without human intervention. Additionally, the low power footprint of our INT4 implementation on the NVIDIA Jetson AGX Orin makes it an ideal candidate for \textbf{Swarm Robotics}, where multiple LiteVLA-powered agents can coordinate in bandwidth-denied environments.

\section{Conclusion}

In this work, we presented LiteVLA-Edge, a deployment-oriented pipeline for fully on-device Vision-Language-Action inference on embedded robotic hardware. By combining supervised image-to-action fine-tuning, post-training 4-bit GGUF quantization, and GPU-accelerated inference through the \texttt{llama.cpp} runtime, we showed that a compact multimodal policy can be integrated into a ROS~2 stack and executed with a mean end-to-end latency of 150.5\,ms, corresponding to approximately 6.6\,Hz under our deployment configuration. Our primary contribution is not a new policy objective or control law, but a practical path for moving compact VLA models from proof-of-concept to reproducible embedded execution. In particular, LiteVLA-Edge preserves modular perception--reasoning--action interfaces, remains fully local, and avoids reliance on cloud infrastructure or desktop-class GPUs. We believe this is a useful systems result because it identifies a realistic deployment point between large general-purpose VLA systems and narrowly reactive controllers. We intentionally scope our conclusions to deployability, timing feasibility, and software integration. Within that scope, our results provide a concrete baseline for fully local, language-conditioned multimodal control on Orin-class platforms. Broader task-level comparisons, longer-duration profiling, and matched evaluations against alternative baselines are natural extensions of the same deployment pipeline, rather than changes to the core method itself. Overall, our results suggest that compact, quantized VLA policies can be packaged into a practical ROS~2-compatible embedded system. As multimodal backbones and efficient runtimes continue to improve, we expect fully local language-conditioned control to become an increasingly practical option for robots operating under bandwidth, power, and latency constraints.
\section{AI Usage Acknowledgement} Chatgpt 5.1 is used to refine the language of this paper.

\bibliographystyle{IEEEtran}

\bibliography{references}

@misc{kim2025openvla,
      title={OpenVLA: An Open-Source Vision-Language-Action Model}, 
      author={Moo Jin Kim and Karl Pertsch and Siddharth Karamcheti and Ted Xiao and Ashwin Balakrishna and Suraj Nair and Rafael Rafailov and Ethan Foster and Grace Lam and Pannag Sanketi and Quan Vuong and Thomas Kollar and Benjamin Burchfiel and Russ Tedrake and Dorsa Sadigh and Sergey Levine and Percy Liang and Chelsea Finn},
      year={2024},
      eprint={2406.09246},
      archivePrefix={arXiv},
      primaryClass={cs.RO},
      url={https://arxiv.org/abs/2406.09246}, 
}

@misc{moondream2_2024,
title        = {Moondream2: A Tiny Vision-Language Model},
  author       = {Verma, Vikhyat and others},
  year         = {2024},
  howpublished = {\url{https://github.com/vikhyat/moondream}},
  note         = {GitHub repository, accessed 2026}
}

@misc{zhou2024tinyllava,
  title        = {TinyLLaVA: A Framework of Small-scale Large Multimodal Models},
  author       = {Zhou, Baichuan and Hu, Ying and Weng, Xi and Jia, Junlong and Luo, Jie and Liu, Xien and Wu, Ji and Huang, Lei},
  year         = {2024},
  eprint       = {2402.14289},
  archivePrefix= {arXiv},
  primaryClass = {cs.LG},
  note         = {arXiv:2402.14289 [cs.LG]}
}

@misc{wang2024qwen2vl,
  title        = {Qwen2-VL: Enhancing Vision-Language Model's Perception of the World at Any Resolution},
  author       = {Wang, Peng and Bai, Shuai and Tan, Sinan and Wang, Shijie and Fan, Zhihao and Bai, Jinze and Chen, Keqin and Liu, Xuejing and Wang, Jialin and Ge, Wenbin and Fan, Yang and Dang, Kai and Du, Mengfei and Ren, Xuancheng and Zhou, Chang and Zhou, Jingren and Lin, Junyang},
  year         = {2024},
  eprint       = {2409.12191},
  archivePrefix= {arXiv},
  primaryClass = {cs.CV},
  note         = {arXiv:2409.12191 [cs.CV]}
}

@misc{steiner2024paligemma2,
  title        = {PaliGemma 2: A Family of Versatile VLMs for Transfer},
  author       = {Steiner, Andreas and Pinto, André Susano and Tschannen, Michael and Keysers, Daniel and Wang, Xiao and Bitton, Yonatan and Gritsenko, Alexey and Minderer, Matthias and Sherbondy, Anthony and Long, Shangbang and Qin, Siyang and Ingle, Reeve and Bugliarello, Emanuele and Kazemzadeh, Sahar and Mesnard, Thomas and Alabdulmohsin, Ibrahim and Beyer, Lucas and Zhai, Xiaohua},
  year         = {2024},
  eprint       = {2412.03555},
  archivePrefix= {arXiv},
  primaryClass = {cs.CV},
  note         = {arXiv:2412.03555 [cs.CV]}
}

@misc{williams2025litevlaefficientvisionlanguageaction,
      title={Lite VLA: Efficient Vision-Language-Action Control on CPU-Bound Edge Robots}, 
      author={Justin Williams and Kishor Datta Gupta and Roy George and Mrinmoy Sarkar},
      year={2025},
      eprint={2511.05642},
      archivePrefix={arXiv},
      primaryClass={cs.RO},
      url={https://arxiv.org/abs/2511.05642}, 
}

@misc{zhang2025edgevla,
    title={EdgeVLA: Efficient Vision-Language-Action Models}, 
      author={Paweł Budzianowski and Wesley Maa and Matthew Freed and Jingxiang Mo and Winston Hsiao and Aaron Xie and Tomasz Młoduchowski and Viraj Tipnis and Benjamin Bolte},
      year={2025},
      eprint={2507.14049},
      archivePrefix={arXiv},
      primaryClass={cs.RO},
      url={https://arxiv.org/abs/2507.14049},
}

@misc{shukor2025smolvla,
 title={SmolVLA: A Vision-Language-Action Model for Affordable and Efficient Robotics}, 
      author={Mustafa Shukor and Dana Aubakirova and Francesco Capuano and Pepijn Kooijmans and Steven Palma and Adil Zouitine and Michel Aractingi and Caroline Pascal and Martino Russi and Andres Marafioti and Simon Alibert and Matthieu Cord and Thomas Wolf and Remi Cadene},
      year={2025},
      eprint={2506.01844},
      archivePrefix={arXiv},
      primaryClass={cs.LG},
      url={https://arxiv.org/abs/2506.01844}, 
}

@inproceedings{zitkovich2023rt2,
  author    = {Anthony Zitkovich and others},
  title     = {{RT-2}: Vision-Language-Action Models Transfer Web Knowledge to Robotic Control},
  booktitle = {Proceedings of the 7th Conference on Robot Learning (CoRL)},
  year      = {2023}
}

@inproceedings{driess2023palme,
  author    = {Danny Driess and others},
  title     = {{PaLM-E}: An Embodied Multimodal Language Model},
  booktitle = {Proceedings of the 40th International Conference on Machine Learning (ICML)},
  year      = {2023}
}

\end{document}